%% file: acl_latex.tex
\documentclass[11pt]{article}

\usepackage[final]{acl}

\usepackage{amssymb}
\usepackage{times}
\usepackage{latexsym}
\usepackage[T1]{fontenc}
\usepackage[utf8]{inputenc}
\usepackage{microtype}
\usepackage{inconsolata}
\usepackage{graphicx}
\usepackage{booktabs}
\usepackage{multirow}
\usepackage{placeins}
\usepackage[most]{tcolorbox}
\usepackage{fancyvrb}
\usepackage{tikz}
\usetikzlibrary{positioning,arrows.meta}

\definecolor{boxgray}{RGB}{90, 90, 90}
\newtcolorbox{prompttemplate}[1]{
  enhanced,
  colback=white,
  colframe=boxgray,
  colbacktitle=boxgray,
  coltitle=white,
  fonttitle=\bfseries,
  title=#1,
  boxrule=0.6pt,
  arc=2pt,
  left=8pt, right=8pt, top=6pt, bottom=6pt,
  breakable,
}

\definecolor{boxblue}{RGB}{22, 60, 135}
\definecolor{boxred}{RGB}{135, 25, 25}
\newtcolorbox{examplebluebox}[1]{
  enhanced, colback=white, colframe=boxblue, colbacktitle=boxblue,
  coltitle=white, fonttitle=\bfseries, title=#1,
  boxrule=0.6pt, arc=2pt, left=8pt, right=8pt, top=6pt, bottom=6pt, breakable,
}
\newtcolorbox{exampleredbox}[1]{
  enhanced, colback=white, colframe=boxred, colbacktitle=boxred,
  coltitle=white, fonttitle=\bfseries, title=#1,
  boxrule=0.6pt, arc=2pt, left=8pt, right=8pt, top=6pt, bottom=6pt, breakable,
}

\title{Vishing-Tactics-Bench: Forecasting Exploitation Trajectories in\\Voice Phishing Calls}

\author{
Jeongmin Lee$^{1,2}$
,
Dongmyung Sul$^{1}$
,
Seung Yun$^{1}$
,
Jinxia Huang$^{1,2\dagger}$
\\
$^{1}$Electronics and Telecommunications Research Institute (ETRI), Korea
\\
$^{2}$University of Science and Technology (UST), Korea
\\
\texttt{\{faraway,dmsul,syun,hgh\}@etri.re.kr}
}

\begin{document}
\maketitle
\footnotetext[2]{Corresponding author.}

\begin{abstract}
Voice phishing (vishing) unfolds in real time; by the time a call has ended and post-hoc
classification is possible, the harm has already been done. The more actionable question is \emph{which concrete harm} (\emph{Information Gathering}
or \emph{Financial Exploitation}) an ongoing call is tactically progressing toward. We present
\textbf{Vishing-Tactics-Bench}, a benchmark grounded in Endsley's situation-awareness (SA) framework that recasts vishing defense from
after-the-fact fraud classification to \emph{harm projection}: predicting at each
turn whether the call will reach either terminal harm. We adapt MITRE ATT\&CK to
vishing as a 6-tactic taxonomy (Vishing-Tactics) and label 35{,}340 scammer utterances
across 5{,}645 synthetic Chinese calls. We define \textbf{Exploitation Trajectory Forecasting},
a survival-style protocol over the two terminal harms with three metrics:
AP@$k$, C-index, and divergence error. Baselines ranging from a Markov heuristic to fine-tuned LLMs
show that the tactical trajectory serves as an interpretable representation of the
call's tactical state, supporting harm-specific
forecasting, which can then be used for the downstream application of
intervention selection;
a stratified lead-time analysis at a tight false-alarm budget further identifies
at what point in a call the trajectory signal yields early
warning.
\end{abstract}

\section{Introduction}

Vishing is a social engineering attack that inflicts financial and informational harm on victims.
Scammers do not deceive at random; they steer the call toward
a strategic goal. A typical vishing call unfolds step by step: contacting the victim, building trust,
applying psychological pressure, and driving the victim toward one of
two terminal harms, \emph{Information Gathering} or \emph{Financial Exploitation}. A vishing call is thus a goal-directed conversation that
unfolds along a sequence of tactics.

{
NLP research has largely framed vishing as call-level binary classification.
But a fraud/benign label, even when accurate, does not specify
\emph{which} concrete harm to block (\emph{Information Gathering} or
\emph{Financial Exploitation}), and a label arriving after the call has
ended is no longer actionable. The actionable defensive question is
instead \textbf{what kind of harm an ongoing call is progressing toward,
and when to intervene}.
}

This work captures that strategic progression at the utterance level. Drawing on the
MITRE ATT\&CK framework from cybersecurity, we build a TTP (Tactics, Techniques,
Procedures) taxonomy adapted to the vishing domain, and assign turn-level tactic labels to
the Chinese synthetic vishing conversations of TeleAntiFraud \citep{ma2025}. Each call is thereby
represented as a sequence of tactics, the call's TTP trajectory.

Building on this trajectory representation, we propose the task of \textbf{Exploitation
Trajectory Forecasting (ETF)}: {a \emph{harm projection} task in which,
at any point in an ongoing call, a model predicts, for each of two terminal harms
(\emph{Information Gathering}, \emph{Financial Exploitation}), whether the call
will reach it within the next $k$ turns}. Unlike static ``fraud / benign''
classification, ETF dynamically models the process by which a strategic
conversation culminates in actual harm. The two harms are
distinct and require different interventions: \emph{Information Gathering}
extracts personal information (itself a harm and a stepping stone), whereas
\emph{Financial Exploitation} is a direct demand for money.

  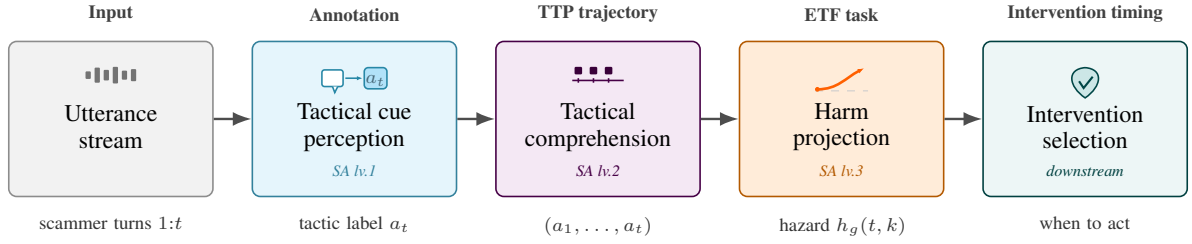
\begin{figure*}[!t]
  \centering
  \begin{tikzpicture}[
    node distance=0.8cm and 0.5cm,
    box/.style={draw, semithick, rounded corners=3pt,
      minimum width=2.7cm, minimum height=2.05cm,
      align=center, font=\small, inner sep=4pt},
    input/.style={box, fill=black!5,   draw=black!45},
    L1/.style=   {box, fill=cyan!9,    draw=cyan!55!black},
    L2/.style=   {box, fill=violet!9,  draw=violet!55!black},
    L3/.style=   {box, fill=orange!15, draw=orange!70!black},
    down/.style= {box, fill=teal!7,    draw=teal!50!black},
    flow/.style ={-Latex, line width=0.9pt, black!70},
    dflow/.style={-Latex, line width=0.9pt, dashed, black!50},
    toplbl/.style={font=\scriptsize\bfseries, color=black!75},
    caplbl/.style={font=\scriptsize, text width=2.7cm, align=center, color=black!80},
    bodytxt/.style={font=\small, align=center},
    salbl/.style ={font=\tiny\itshape},
  ]
    \node[input]              (utt)  {};
    \node[L1,   right=of utt] (per)  {};
    \node[L2,   right=of per] (com)  {};
    \node[L3,   right=of com] (proj) {};
    \node[down, right=of proj](int)  {};

    \node[anchor=north] at ([yshift=-6pt]utt.north) {%
      \tikz[baseline]{%
        \foreach \x/\h in {-0.30/0.10,-0.18/0.20,-0.06/0.14,0.06/0.22,0.18/0.12,0.30/0.16}
          \fill[black!55, rounded corners=0.4pt] (\x,-\h/2) rectangle ++(0.07,\h);}};
    \node[bodytxt] at ([yshift=-3pt]utt.center) {Utterance\\stream};

    \node[anchor=north] at ([yshift=-6pt]per.north) {%
      \tikz[baseline]{%
        \draw[fill=white, draw=cyan!60!black, semithick, rounded corners=1pt]
          (-0.42,-0.08) rectangle (-0.14,0.14);
        \draw[fill=white, draw=cyan!60!black, semithick]
          (-0.28,-0.08) -- (-0.25,-0.16) -- (-0.21,-0.08);
        \draw[cyan!60!black, semithick, -{Latex[length=2.2pt]}]
          (-0.11,0.03) -- (0.12,0.03);
        \draw[fill=cyan!28, draw=cyan!60!black, semithick, rounded corners=1.5pt]
          (0.14,-0.11) rectangle (0.44,0.17);
        \node[font=\scriptsize\bfseries, color=cyan!35!black, inner sep=0pt]
          at (0.29,0.08) {$a_t$};}};
    \node[bodytxt] at ([yshift=-3pt]per.center) {Tactical cue\\perception};
    \node[salbl, anchor=south, color=cyan!55!black]
          at ([yshift=4pt]per.south)  {SA lv.1};

    \node[anchor=north] at ([yshift=-6pt]com.north) {%
      \tikz[baseline]{%
        \draw[violet!55!black, semithick] (-0.34,-0.08) -- (0.34,-0.08);
        \foreach \x in {-0.26,-0.06,0.14} {
          \draw[violet!55!black, semithick] (\x,-0.11) -- (\x,-0.05);
          \fill[violet!45!black, rounded corners=0.5pt]
            (\x-0.05, 0.00) rectangle (\x+0.05, 0.10);
        }}};
    \node[bodytxt] at ([yshift=-3pt]com.center) {Tactical\\comprehension};
    \node[salbl, anchor=south, color=violet!55!black]
          at ([yshift=4pt]com.south)  {SA lv.2};

    \node[anchor=north] at ([yshift=-6pt]proj.north) {%
      \tikz[baseline]{%
        \draw[gray!40, dashed] (-0.35,-0.12) -- (0.35,-0.12);
        \draw[orange!75!red, line width=1pt, -{Latex[length=2.5pt]}]
          (-0.32,-0.10) .. controls (-0.05,-0.10) and (0.05,0.05) .. (0.30,0.18);
        \fill[orange!75!red] (-0.32,-0.10) circle (0.9pt);}};
    \node[bodytxt] at ([yshift=-3pt]proj.center) {Harm\\projection};
    \node[salbl, anchor=south, color=orange!70!black]
          at ([yshift=4pt]proj.south) {SA lv.3};

    \node[anchor=north] at ([yshift=-6pt]int.north) {%
      \tikz[baseline]{%
        \draw[fill=teal!20, draw=teal!55!black, semithick]
          (0,0.20) .. controls (0.25,0.18) and (0.22,-0.05) ..
          (0,-0.20) .. controls (-0.22,-0.05) and (-0.25,0.18) .. (0,0.20);
        \draw[teal!55!black, line width=0.8pt]
          (-0.08,0.02) -- (-0.02,-0.07) -- (0.10,0.08);}};
    \node[bodytxt] at ([yshift=-3pt]int.center) {Intervention\\selection};
    \node[salbl, anchor=south, color=teal!50!black]
          at ([yshift=4pt]int.south)  {downstream};

    \draw[flow]  (utt)  -- (per);
    \draw[flow]  (per)  -- (com);
    \draw[flow]  (com)  -- (proj);
    \draw[flow] (proj) -- (int);

    \node[toplbl, above=0.12cm of utt]  {Input};
    \node[toplbl, above=0.12cm of per]  {Annotation};
    \node[toplbl, above=0.12cm of com]  {TTP trajectory};
    \node[toplbl, above=0.12cm of proj] {ETF task};
    \node[toplbl, above=0.12cm of int]  {Intervention timing};

    \node[caplbl, below=0.12cm of utt]  {scammer turns $1{:}t$};
    \node[caplbl, below=0.12cm of per]  {tactic label $a_t$};
    \node[caplbl, below=0.12cm of com]  {$(a_1,\dots,a_t)$};
    \node[caplbl, below=0.12cm of proj] {hazard $h_g(t,k)$};
    \node[caplbl, below=0.12cm of int]  {when to act};
  \end{tikzpicture}
  \caption{From utterances to intervention timing,
  following Endsley's situation-awareness (SA) framework.
  Vishing-Tactics-Bench implements lv.1 (turn-level TTP labeling), lv.2
  (TTP trajectory), and lv.3 (ETF task).
  Trajectory dynamics are analyzed in
  Figure~\ref{fig:leadbreak} and Table~\ref{tab:outcome-hazard}.}
  \label{fig:sa_pipeline}
  \end{figure*}

{
Cognitively, this view aligns with the situation-awareness (SA) framework
\citep{endsley1995}, which decomposes operator awareness into
\emph{perception} of cues, \emph{comprehension} of their meaning,
and \emph{projection} of how the situation unfolds.
Vishing-Tactics-Bench instantiates the three SA levels as
turn-level TTP labeling (perception), the TTP trajectory
(comprehension), and Exploitation Trajectory Forecasting (projection;
Figure~\ref{fig:sa_pipeline}).
}

\vspace{1.5em}

Our contributions are as follows.

\begin{enumerate}
\item \textbf{A turn-level TTP-annotated corpus.} We adapt MITRE ATT\&CK to
vishing as a 6-tactic taxonomy (Vishing-Tactics) and apply it to 5{,}645 Chinese vishing calls
from TeleAntiFraud, assigning tactic labels to every one of their 35{,}340 scammer
utterances, the first utterance-level tactical structure on this data, which
previously carried only call-level labels \citep{ma2025,wang2026}.
\item {\textbf{Exploitation Trajectory Forecasting (ETF) as harm
projection.} Inspired by survival-style time-to-event evaluation,
we define ETF as predicting, at each turn, the hazard that each of two terminal
harms is reached within $k$ turns, and adopt a turn-level
evaluation protocol (AP@$k$, C-index, divergence error).}
\item \textbf{Baselines and analysis.} With baselines ranging from a Markov
heuristic to encoder and fine-tuned LLM models, we analyze the difficulty and
structure of ETF and show that the tactical trajectory is a key predictive signal
whose value depends on how it is integrated into the model.
\end{enumerate}

\section{Related Work}

\paragraph{Vishing and scam-call NLP.}
The literature on scam-call NLP falls into two lines. The dominant line
is \emph{call-level classification}, which assigns a single label per
completed call: TeleAntiFraud-28k \citep{ma2025}, SAFE-QAQ \citep{wang2026},
Korean vishing detection \citep{boussougou2023}, and LLM-augmented
expert-rule detection under limited labels \citep{ma2025b}. A second, emerging line moves below the call level toward in-call,
real-time, deployment-aware detection
\citep{shen2025,ocleicniks2025real,cho2025towards} and psychological-technique annotation of real-world
scam reports \citep{psyscam2025}; yet its predictions
remain classifications of the current state rather than forecasts of which
terminal harm a call is heading toward. Field studies confirm
that vishing reliably succeeds against real users \citep{tu2019}, yet
existing benchmarks measure post-hoc classification rather than live-call
forecasting. ETF fills the resulting gap by predicting, at the turn level,
which terminal harm an ongoing call is heading toward.
Call-level detection settles the upstream
question of whether a call is vishing at all; ETF operates downstream,
on calls that decision has already flagged.

\paragraph{Attack-stage modelling in cybersecurity.}
Modelling adversarial behaviour as an ordered sequence of tactical stages
has a long history in security. The Cyber Kill Chain \citep{hutchins2011}
formalises intrusion as a seven-stage process, and MITRE ATT\&CK
\citep{strom2018} generalises this into a tactic--technique--procedure
ontology, which is now standard across threat intelligence. \citet{montanez2022}
adapt the same framework to social engineering as a Cyber Social
Engineering Kill Chain, and \citet{tsinganos2023} apply dialogue state
tracking to recognise Cyber Social Engineering (CSE) attacks turn by turn. Closer to our setting,
\citet{wood2023analysis} extract scam stages and scripts from scam-baiting calls
using topic modelling and HMM-based stage transitions. Our Vishing-Tactics taxonomy inherits the
tactic/technique abstraction from ATT\&CK and the social-engineering /
scam-stage focus from \citet{montanez2022} and \citet{wood2023analysis}, but where prior work
\emph{recognises} or \emph{extracts} the current stage, we use
the trajectory of stages as the input to a forecasting model.

\paragraph{Survival analysis and conversational forecasting.}
The forecasting protocol underlying ETF stands at the intersection of two
lines. The first is \emph{survival analysis}, whose foundations are the
proportional-hazards model \citep{cox1972}, the Kaplan--Meier
product-limit estimator \citep{kaplan1958}, and the concordance index
\citep{harrell1982}, from which we borrow time-to-event
terminology and rank-based evaluation. The second line
is \emph{conversational forecasting}: dialogue-side work has predicted
conversational outcomes (derailment \citep{zhang2018,altarawneh2023},
negotiation success \citep{sicilia2024}, and chat-based social-engineering
intent \citep{convosentinel2024}), always as a single binary outcome
rather than a hazard trajectory. ETF is not intended as a
clinical-style survival-modeling benchmark; rather, it formulates a
streaming, harm-specific forecasting problem on a tactical
trajectory of an adversarial dialogue.

\paragraph{Cognitive and psychological framing.}
The framing of ETF as harm \emph{projection} draws on three psychological
threads. First, the situation-awareness model of \citet{endsley1995}
decomposes operator awareness into perception, comprehension, and
projection; ETF corresponds to the projection level, built upon perception
(tactic labelling) and comprehension (the trajectory). Second, the
scammer side of the call has been analysed extensively in compliance
psychology \citep{cialdini2018influence}, and recent NLP corpora annotate the
manipulation techniques themselves
\citep{wang2024,psyscam2025}. These provide the conceptual basis for the
trust-building / pressure / information-gathering distinctions that our
Vishing-Tactics taxonomy formalises. Third, qualitative stage models of specific
scam genres \citep{whitty2013} characterise the scammer's strategic
progression in a complementary, narrative form. Together these accounts
motivate \emph{why} the tactical trajectory matters (each tactic
activates a specific compliance mechanism), and locate harm projection
as the cognitive level at which a defender can intervene meaningfully.

\section{Vishing-Tactics Taxonomy}

Following MITRE ATT\&CK \citep{strom2018}, we build a vishing TTP taxonomy.
A \emph{tactic} is the attacker's objective, a \emph{technique} the concrete
method, and a \emph{procedure} the detailed process --- TTP thus describes an
attack at the level of the \emph{attacker's intent and plan} rather than its
surface means.

The Vishing-Tactics taxonomy consists of six tactics:

\begin{itemize}
\itemsep0pt
\item \textbf{T1001 Initial Contact} --- making first contact with the victim.
\item \textbf{T1002 Trust Building} --- impersonating an authority or building rapport.
\item \textbf{T1003 Psychological Pressure} --- inducing fear, urgency, or a benefit lure.
\item \textbf{T1004 Information Gathering} --- eliciting personal/financial credentials.
\item \textbf{T1005 Financial Exploitation} --- directing the victim to move money.
\item \textbf{T1006 Evasion} --- isolating the victim from family, bank, or police.
\end{itemize}

Utterances with no tactical intent are labeled NONE. Full definitions and
distributions are in Appendix~\ref{sec:appendix-a}.
The benchmark uses tactic-level labels throughout;
technique-level annotation is released for future fine-grained analysis.

Of these six, \emph{Information Gathering} (T1004) and
\emph{Financial Exploitation} (T1005) are the \emph{terminal harms} ETF
forecasts; the rest are the process driving the call toward them.

\section{Dataset Construction}
\label{sec:dataset}

\paragraph{Base data and tactic annotation.}
Our corpus is based on the Chinese synthetic vishing conversations of TeleAntiFraud-28k
\citep{ma2025}, generated by an LLM with synthesized TTS audio. Fraud types include
customer-service impersonation, bank, investment, phishing, lottery, and kidnap; the
fraud type of each call follows the original TeleAntiFraud label. We assigned the
tactic labels of Section~3 to scammer utterances; T1001
(Initial Contact) does not apply to mid-call recordings, so annotation
covers the remaining five tactics (T1002--T1006) and their 21 techniques,
plus NONE. Two LLMs (Claude and GPT-4o)
independently labeled every utterance after viewing the full call context
(inter-model Cohen's $\kappa = 0.94$). Two human
annotators relabeled a stratified $\sim$10\% sample of the test split
(inter-human $\kappa = 0.69$; joint Fleiss's $\kappa = 0.66$). Details
in Appendix~\ref{sec:appendix-a}.

\paragraph{Standard split.}
For ETF, we sampled the 5{,}645 vishing calls (out of TeleAntiFraud-28k) for
which both a tactic label and the original fraud-type label are available,
and stratified the split by fraud type (Table~\ref{tab:split-overall}; an
average of 6.3 scammer turns per call).
\begin{table}[h]
\centering
\small
\begin{tabular}{lrrr}
\toprule
\textbf{Unit} & \textbf{Total} & \textbf{Train} & \textbf{Test} \\
\midrule
Calls         &  5{,}645 &  4{,}515 &  1{,}130 \\
Scammer turns & 35{,}340 & 28{,}222 &  7{,}118 \\
Victim turns  & 28{,}631 & 22{,}913 &  5{,}718 \\
Total turns   & 63{,}971 & 51{,}135 & 12{,}836 \\
\bottomrule
\end{tabular}
\caption{Vishing-Tactics-Bench standard split totals (per-fraud-type
breakdown in Appendix~\ref{sec:appendix-b}, Table~\ref{tab:split}).}
\label{tab:split-overall}
\end{table}

\paragraph{Tactic distribution.}
Tactic counts over the standard split span an order of magnitude
(Table~\ref{tab:tactic-dist}). Trust Building, Psychological Pressure, and
Information Gathering together account for $82.2\%$ of scammer turns, while the
terminal-stage tactics \emph{Financial Exploitation} and \emph{Evasion}
are each near $1.4\%$, and NONE (no tactical intent) at $15.0\%$. This
$\sim$$20{\times}$ frequency gap between \emph{Information Gathering}
($31.5\%$) and \emph{Financial Exploitation} ($1.4\%$) is the
turn-level imprint of the difficulty contrast ETF must handle within
one task.
\begin{table}[h]
\centering
\small
\begin{tabular}{lrr}
\toprule
\textbf{Tactic} & \textbf{Count} & \textbf{\%} \\
\midrule
T1002 Trust Building          &  8{,}384 & 23.7 \\
T1003 Psychological Pressure  &  9{,}554 & 27.0 \\
T1004 Information Gathering   & 11{,}133 & 31.5 \\
T1005 Financial Exploitation  &     481 &  1.4 \\
T1006 Evasion                 &     496 &  1.4 \\
NONE                          &  5{,}292 & 15.0 \\
\midrule
\textbf{Total}                & \textbf{35{,}340} & \textbf{100.0} \\
\bottomrule
\end{tabular}
\caption{Tactic-level distribution over the Vishing-Tactics-Bench standard
split.}
\label{tab:tactic-dist}
\end{table}

\paragraph{Distribution of the two terminal harms.}
The fraud type determines the harm pathway: kidnap fraud goes almost
directly to \emph{Financial Exploitation} (28.7\%) with little
\emph{Information Gathering} (5.7\%), whereas the other types have high
\emph{Information Gathering} reach (69.0--98.6\%); phishing
almost entirely terminates at \emph{Information Gathering} (98.6\%). The
two harms are not in a linear relation: about 36\% of calls
reaching \emph{Financial Exploitation} do so without passing through
\emph{Information Gathering} (full corpus, $n{=}365$ reaching calls).
Overall, \emph{Information Gathering} is common (85.8\%) while
\emph{Financial Exploitation} is rare (6.4\%). Per-type reach rates are
in Appendix~\ref{sec:appendix-b}, Table~\ref{tab:reach}.

\section{Task: Exploitation Trajectory Forecasting}
\label{sec:task}

Given the TTP trajectory up to turn $t$,
ETF predicts the hazard of reaching each terminal harm within the next $k$ turns.
ETF conditions on a call already identified as vishing;
separating vishing from benign calls is an upstream detection task.

\paragraph{Definition.}
We view a vishing call as a sequence of scammer utterances. The $t$-th scammer utterance
carries a tactic $a_t$, and the full sequence $(a_1, \ldots, a_T)$ is the call's
TTP trajectory. ETF predicts the occurrence of the two terminal harms
$g \in \{\textrm{T1004}, \textrm{T1005}\}$.
The first appearance of $g$ marks the onset of the
corresponding victim-facing harm: information disclosure for T1004,
financial loss for T1005. Let $\tau_g$ be the scammer turn at which
tactic $g$ first appears ($\tau_g = \infty$ if it never appears). At each scammer turn
$t$, conditioned on $g$ not yet reached, the model predicts the \emph{hazard}
\begin{equation}
\label{eq:hazard}
h_g(t, k) = \Pr[\, \tau_g \le t + k \mid X_t,\ \tau_g > t \,]
\end{equation}
of reaching $g$ within the next $k$ turns, for each $g$ and $k \in \{1, 3, 5\}$.
$X_t$ is the observation up to turn $t$. A single model outputs both harms' hazards.

\paragraph{Input.}
$X_t$ has two configurations: (i) the dialogue utterances up to turn $t$, and (ii)
the utterances plus the TTP trajectory $(a_1, \ldots, a_t)$. Comparing (ii) against
(i) measures the contribution of tactical information.

\paragraph{Evaluation.}
ETF uses three metrics, computed separately for the two terminal harms.

\begin{itemize}
\itemsep2pt
\item \textbf{AP@$k$} --- average precision with ground truth ``reaches $g$ within the
next $k$ turns'' and the hazard $h_g(t,k)$ as score. It is suited to the rare
positives of \emph{Financial Exploitation}, for which accuracy is uninformative.
Higher is better.
\item \textbf{C-index} --- the concordance index
\citep{harrell1982,zhao2021bertsurv}: the fraction of (reaching, non-reaching)
call pairs in which the model assigns the higher hazard to the reaching call.
0.5 is random, higher is better.
\item \textbf{Divergence error} --- a task-specific metric: the distance, in
turns, between the turn at which the predicted hazard curve rises most sharply
and the turn at which $g$ first appears.
Let $\hat{h}_g(t,k)$ be the predicted hazard at turn $t$ for
horizon $k$. The predicted and ground-truth branch turns are
{\small
\begin{gather}
\hat{\tau}_g = 1 + \arg\max_{i}\big[\hat{h}_g(t_{i+1},k) - \hat{h}_g(t_i,k)\big], \label{eq:pred-branch}\\
\tau_g = \min\{\,i : a_i = g\,\} \label{eq:gt-branch}
\end{gather}
}
and the divergence error averages their absolute distance over the calls that
reach $g$:
\begin{equation}
\mathrm{Div}_g \;=\; \mathbb{E}\big[\,|\hat{\tau}_g - \tau_g|\;\big|\;g\text{ reached}\big].
\label{eq:diverr}
\end{equation}
We use $k=3$. Calls that never reach $g$ have no ground-truth branch and are
excluded from the average.
Its range is $[0,\text{call length})$ (mean 6.3 turns); lower is better.

\end{itemize}

\section{Baselines}

We set up baselines of differing character,
spanning statistical, discriminative, and generative modeling
families.

\begin{itemize}
\itemsep2pt
\item \textbf{Markov heuristic.} Estimates the tactic transition matrix
$P(T_{t+1} \mid T_t)$ from training trajectories and iterates it to compute reach
probabilities. It uses no dialogue text and shows how far ETF is predictable from the
transition structure alone.
\item \textbf{Encoder-based learned model.} A learned hazard head over both terminal
harms, with three input configurations: \textbf{text only} (a Chinese pretrained
encoder, \texttt{chinese-roberta-wwm-ext} \citep{cui2021chinesebert}), \textbf{TTP
only} (a small GRU over the tactic-label trajectory; LSTM and
Transformer variants in Appendix~\ref{sec:appendix-seqenc}), and \textbf{text+TTP} (both
encoders, late-fused by concatenation).
\item \textbf{LLM zero-shot.} Prompts gpt-4.1-mini to
predict reach directly, under prompts with and without the TTP
trajectory.
\item \textbf{LLM fine-tuning.} Supervised-fine-tunes two LLMs
(gpt-4.1-mini via the OpenAI tuning API; Qwen2.5-7B \citep{qwen2025technical} via LoRA,
$r{=}16$) on the train split, with and without the TTP.
\end{itemize}

The TTP-included/excluded comparison is the central analysis axis, directly testing
whether tactical information provides a signal beyond text.

\section{ETF Results}
\label{sec:results}

The general pattern in Table~\ref{tab:results} is consistent across both
terminal harms: learned baselines (encoder, supervised-fine-tuned LLMs)
substantially outperform non-learned baselines (the Markov heuristic
and zero-shot gpt-4.1-mini), with the gap most pronounced on
\emph{Information Gathering}.
Each TTP-using model is reported twice---\emph{gold}
(oracle TTP) and \emph{pred} (cascaded classifier)---so the
deployment-relevant noise can be read off directly; text-only rows are
unaffected by the TTP source.
The gold-versus-predicted comparison quantifies how
upstream TTP recognition errors propagate into downstream harm
forecasting.

\begin{table*}[t]
\centering
\small
\setlength{\tabcolsep}{5pt}
\begin{tabular}{llcccccc}
\toprule
& & \multicolumn{3}{c}{\textbf{\emph{Information Gathering}}}
& \multicolumn{3}{c}{\textbf{\emph{Financial Exploitation}}} \\
\cmidrule(lr){3-5}\cmidrule(lr){6-8}
\textbf{Model} & \textbf{TTP} & \textbf{AP@1/3/5}\,$\uparrow$ & \textbf{C-idx}\,$\uparrow$
& \textbf{Div.\,err.}\,$\downarrow$ & \textbf{AP@1/3/5}\,$\uparrow$
& \textbf{C-idx}\,$\uparrow$ & \textbf{Div.\,err.}\,$\downarrow$ \\
\midrule
Base rate        & --- & 0.27 / 0.45 / 0.48 & 0.50 & n/a & 0.02 / 0.03 / 0.04 & 0.50 & n/a \\
\midrule
\multirow{2}{*}{Markov heuristic} & gold & 0.40 / 0.49 / 0.52 & 0.55 & 1.45 & 0.39 / 0.24 / 0.22 & 0.67 & 0.07 \\
                                  & pred & 0.39 / 0.51 / 0.53 & 0.58 & 1.51 & 0.19 / 0.15 / 0.14 & 0.65 & 0.11 \\
\midrule
Encoder (text)              & ---  & 0.89 / 0.95 / \textbf{0.97} & \textbf{0.98} & 1.43 & 0.58 / 0.56 / 0.56 & \textbf{0.92} & 1.15 \\
\multirow{2}{*}{Encoder (TTP)}      & gold & 0.32 / 0.71 / 0.79 & 0.84 & 2.10 & 0.11 / 0.10 / 0.10 & 0.68 & 0.89 \\
                                    & pred & 0.32 / 0.69 / 0.77 & 0.83 & 2.12 & 0.08 / 0.09 / 0.10 & 0.68 & 1.45 \\
\multirow{2}{*}{Encoder (text+TTP)} & gold & 0.89 / 0.95 / \textbf{0.97} & \textbf{0.98} & 1.36 & 0.59 / 0.57 / 0.56 & 0.91 & 1.15 \\
                                    & pred & 0.89 / 0.95 / \textbf{0.97} & \textbf{0.98} & 1.38 & 0.60 / \textbf{0.58} / \textbf{0.57} & \textbf{0.92} & 1.25 \\
\midrule
Qwen2.5-7B SFT (text)               & --- & 0.54 / 0.73 / 0.80 & 0.86 & 1.51 & 0.09 / 0.08 / 0.08 & 0.54 & 1.96 \\
\multirow{2}{*}{Qwen2.5-7B SFT (text+TTP)} & gold & 0.67 / 0.81 / 0.88 & 0.93 & 1.48 & 0.32 / 0.20 / 0.18 & 0.60 & 0.18 \\
                                          & pred & 0.60 / 0.79 / 0.85 & 0.91 & 1.45 & 0.16 / 0.12 / 0.11 & 0.60 & 0.25 \\
\midrule
gpt-4o ZS (text) & --- & 0.28 / 0.37 / 0.39 & 0.31 & 0.76 & 0.11 / 0.11 / 0.10 & 0.64 & 0.96 \\
\multirow{2}{*}{gpt-4o ZS (text+TTP)} & gold & 0.34 / 0.41 / 0.42 & 0.40 & \textbf{0.53} & 0.26 / 0.18 / 0.15 & 0.65 & 0.52 \\
 & pred & 0.33 / 0.40 / 0.42 & 0.40 & 0.61 & 0.17 / 0.14 / 0.13 & 0.65 & 0.38 \\
\midrule
gpt-4.1-mini ZS (text)              & --- & 0.27 / 0.39 / 0.41 & 0.39 & 1.22 & 0.07 / 0.09 / 0.09 & 0.65 & 1.49 \\
\multirow{2}{*}{gpt-4.1-mini ZS (text+TTP)} & gold & 0.36 / 0.43 / 0.45 & 0.45 & 0.58 & 0.35 / 0.22 / 0.18 & 0.65 & 0.34 \\
                                            & pred & 0.35 / 0.43 / 0.45 & 0.46 & 0.59 & 0.22 / 0.17 / 0.14 & 0.66 & 0.45 \\
gpt-4.1-mini SFT (text)             & --- & 0.87 / 0.91 / 0.93 & 0.96 & 1.24 & 0.51 / 0.46 / 0.43 & 0.83 & 0.76 \\
\multirow{2}{*}{gpt-4.1-mini SFT (text+TTP)} & gold & \textbf{0.93} / \textbf{0.96} / \textbf{0.97} & \textbf{0.98} & 1.28 & \textbf{0.67} / 0.55 / 0.53 & 0.86 & \textbf{0.00} \\
                                             & pred & 0.85 / 0.91 / 0.92 & 0.95 & 1.25 & 0.30 / 0.31 / 0.33 & 0.80 & 0.14 \\
\bottomrule
\end{tabular}
\caption{ETF results on the standard test split. Arrows
($\uparrow$/$\downarrow$): better direction. AP@$k$, C-index $\in [0, 1]$;
divergence error in turns. The \textbf{TTP} column denotes the TTP source:
\emph{gold} (LLM-annotated oracle) vs.\ \emph{pred} (cascaded RoBERTa
turn-level classifier; macro-F1 0.75). \emph{Text}-only rows do not use
TTP (---). Encoder (3 variants) and Qwen2.5-7B SFT (2 variants) are mean
across 3 random seeds; other rows are single runs (3-seed std and 95\%
bootstrap CI in Appendix
Tables~\ref{tab:results-uncertainty-t1004}--\ref{tab:results-uncertainty-t1005}).
\textbf{Bold}: best per column.
}%
\label{tab:results}
\end{table*}

\paragraph{AP@$k$.} Learning closes the zero-shot--learned
gap by roughly an order of magnitude. \emph{Information Gathering} is
recovered quickly (ZS 0.27--0.45, learned 0.87--0.97). \emph{Financial
Exploitation} stays lower in absolute value (window base rate 0.02--0.04)
but rises by the same relative factor (ZS 0.07--0.39, learned 0.51--0.67);
FE AP must be read against the base-rate row (e.g., Markov FE 0.39
$\approx$ 20$\times$ base rate; Markov IG 0.40 is only marginally above
chance).
\paragraph{C-index.} Text-only zero-shot gpt-4.1-mini is 0.39 for
\emph{Information Gathering} (below random), and adding the TTP
trajectory does not repair this; the encoder and gpt-4.1-mini SFT
reach 0.96--0.98.
\emph{Financial Exploitation} C-index improves modestly (Markov 0.67 to
encoder 0.92), since Markov's transitions already rank usefully.

\paragraph{Divergence error.} For \emph{Financial Exploitation},
several transition- or trajectory-aware rows locate the branch very
close to the true turn, including Markov, Qwen2.5-7B SFT (text+TTP),
and gpt-4.1-mini SFT (text+TTP), while text-only and TTP-only neural
variants are less precise. For \emph{Information Gathering}, most
models fall in the 1.2--2.1 range; gpt-4.1-mini ZS (text+TTP) is the
lone outlier near 0.6. T1004 occurs gradually across the call, so
there is no sharp branch for any model to locate.

\paragraph{TTP carries non-redundant signal, but its gain depends on
integration.}
Encoder (TTP) alone---without dialogue text---attains
\emph{Information Gathering} AP@3 0.71 and C-index 0.84, surpassing every
text-using zero-shot LLM, and adding the trajectory raises gpt-4.1-mini's
\emph{Financial Exploitation} AP@1 fivefold (0.07 to 0.35). With oracle
TTP, the largest text+TTP gains on \emph{Financial Exploitation} AP@1 go
to the inline-fusion LLMs (gpt-4.1-mini SFT $+0.16$, Qwen2.5-7B SFT
$+0.23$), while the encoder gains marginally ($+0.01$) because text alone
already saturates \emph{Information Gathering} (text-only C-index 0.98).
The two integration methods differ: the encoder concatenates a small TTP
branch with a much larger text representation (late fusion), while LLM
SFT reads TTP tokens inline within the same attention context.
Zero-shot GPT-4o shows the same trend: the trajectory
raises \emph{Financial Exploitation} AP@1 from 0.11 to 0.26 with oracle
TTP and 0.17 with predicted TTP, still well below the encoder's 0.60.

\paragraph{Late fusion is robust to predicted-TTP noise; inline fusion is not.}
Substituting oracle TTP with the cascaded classifier (macro-F1
0.75) preserves the encoder's \emph{Financial Exploitation} AP@1 (gold
0.59, pred 0.60) but cuts it to roughly half for the inline-fusion LLMs
(gpt-4.1-mini SFT 0.67 $\to$ 0.30; Qwen2.5-7B SFT 0.32 $\to$ 0.16). The
contrast is asymmetric across the two harms: gpt-4.1-mini SFT loses only
$0.08$ AP@1 on \emph{Information Gathering} ($0.93\to0.85$) but $0.37$ on
\emph{Financial Exploitation} ($0.67\to0.30$), tracking the classifier's
own per-class F1 of 0.90 on T1004 vs.\ 0.58 on T1005. Rare-tactic
classification noise thus propagates into rare-harm prediction when TTP
is read inline. In the deployment-relevant predicted-TTP setting the
encoder (text+TTP) is the strongest model on \emph{Financial Exploitation}
AP@1 (0.60),
surpassing gpt-4.1-mini SFT (0.30) by a factor of two.

\paragraph{The two terminal harms differ contrastively.}
\emph{Information Gathering} is common (85.8\% reach)
with relatively gradual unfolding, while \emph{Financial Exploitation}
is rare (6.4\%) but appears abruptly and is timed precisely. The two
harms differ along opposite axes (frequency vs.\ timing), and ETF reports
both. The TTP gain is also \emph{type-conditional}: largest on kidnap
($+0.45$ \emph{IG} AP@1, gpt-4.1-mini SFT) and investment ($+0.31$
\emph{FE} AP@1), but slightly negative on bank/lottery; per-type
analysis appears in Appendix~\ref{sec:per-fraud-full}. Per-type
results on kidnap ($n{=}46$) and lottery ($n{=}60$) use small test sets and should
be read as suggestive.

\section{Anatomy of the Two Terminal Harms}
\label{sec:anatomy}

\begin{table*}[t]
\centering
\small
\setlength{\tabcolsep}{5pt}
\begin{tabular}{llccccccc}
\toprule
& & \multicolumn{4}{c}{\textbf{\shortstack{\emph{Financial Exploitation} hazard@3\\by call outcome}}}
& \multicolumn{3}{c}{\textbf{\shortstack{Both calls: \emph{FE} hazard@3\\vs.\ the \emph{IG} branch}}} \\
\cmidrule(lr){3-6}\cmidrule(lr){7-9}
\textbf{Model} & \textbf{TTP} & \textbf{Both}\,$\uparrow$ & \textbf{\emph{FE}-only}\,$\uparrow$ & \textbf{\emph{IG}-only}\,$\downarrow$ & \textbf{Neither}\,$\downarrow$
& \textbf{Before} & \textbf{After} & \textbf{$\Delta$}\,$\uparrow$ \\
\midrule
\multirow{2}{*}{Markov heuristic} & gold & 0.086 & 0.097 & 0.048 & 0.048 & 0.050 & 0.047 & $-0.003$ \\
                                  & pred & 0.091 & 0.118 & 0.050 & 0.058 & 0.050 & 0.050 & $0.000$ \\
\midrule
Encoder (text)               & ---  & 0.253 & 0.330 & 0.011 & 0.063 & 0.238 & 0.276 & $+0.039$ \\
\multirow{2}{*}{Encoder (TTP)}      & gold & 0.052 & 0.100 & 0.044 & 0.066 & 0.084 & 0.024 & $-0.059$ \\
                                    & pred & 0.056 & 0.117 & 0.044 & 0.073 & 0.082 & 0.029 & $-0.053$ \\
\multirow{2}{*}{Encoder (text+TTP)} & gold & 0.282 & 0.326 & 0.014 & 0.059 & 0.249 & 0.365 & $+0.116$ \\
                                    & pred & 0.262 & 0.312 & 0.011 & 0.055 & 0.229 & 0.318 & $+0.089$ \\
\midrule
Qwen2.5-7B SFT (text)               & ---  & 0.070 & 0.142 & 0.001 & 0.018 & 0.000 & 0.009 & $+0.009$ \\
\multirow{2}{*}{Qwen2.5-7B SFT (text+TTP)} & gold & 0.161 & 0.186 & \textbf{0.000} & \textbf{0.005} & 0.000 & 0.000 & $0.000$ \\
                                          & pred & 0.169 & 0.261 & 0.008 & 0.031 & 0.000 & 0.009 & $+0.009$ \\
\midrule
gpt-4o ZS (text) & ---  & 0.423 & 0.525 & 0.299 & 0.290 & 0.237 & 0.383 & $+0.146$ \\
\multirow{2}{*}{gpt-4o ZS (text+TTP)} & gold & 0.492 & 0.551 & 0.330 & 0.249 & 0.249 & 0.458 & $+0.208$ \\
 & pred & \textbf{0.502} & \textbf{0.591} & 0.337 & 0.302 & 0.255 & 0.463 & $+0.208$ \\
\midrule
gpt-4.1-mini ZS (text)              & ---  & 0.274 & 0.417 & 0.166 & 0.184 & 0.146 & 0.202 & $+0.056$ \\
\multirow{2}{*}{gpt-4.1-mini ZS (text+TTP)} & gold & 0.410 & 0.455 & 0.241 & 0.167 & 0.139 & 0.370 & $\mathbf{+0.231}$ \\
                                            & pred & 0.401 & 0.502 & 0.235 & 0.231 & 0.161 & 0.321 & $+0.159$ \\
gpt-4.1-mini SFT (text)             & ---  & 0.118 & 0.285 & 0.011 & 0.068 & 0.048 & 0.043 & $-0.006$ \\
\multirow{2}{*}{gpt-4.1-mini SFT (text+TTP)} & gold & 0.149 & 0.279 & 0.008 & 0.047 & 0.031 & 0.021 & $-0.010$ \\
                                             & pred & 0.136 & 0.260 & 0.015 & 0.074 & 0.031 & 0.032 & $+0.002$ \\
\bottomrule
\end{tabular}

\caption{Anatomy of \emph{Financial Exploitation} hazard; \textbf{TTP}
source as in Table~\ref{tab:results}. \textbf{(Left)} mean predicted
\emph{Financial Exploitation} hazard@3 by call outcome.
\textbf{(Right)} for Both calls, mean \emph{Financial Exploitation}
hazard@3 before vs.\ after the \emph{Information Gathering} branch,
restricted to turns prior to \emph{Financial Exploitation}; a model that
updates dynamically yields $\Delta>0$.}
\label{tab:outcome-hazard}
\end{table*}

Table~\ref{tab:results} pools across calls and turns; here we condition
on each call's \emph{terminal outcome} to keep
opposite behaviours on outcome groups from cancelling.

\paragraph{Call outcome typology.}
The two harms are neither strictly sequential
nor independent (typology in Appendix
Table~\ref{tab:quadrant}): $\sim$5\% of \emph{Information Gathering}-reaching
calls reach \emph{Financial Exploitation}, while $\sim$40\% of
\emph{Financial Exploitation}-reaching calls bypass \emph{Information
Gathering} ($n{=}67$ in test). Neither-outcome calls are short
(median 2 turns) and stall in trust building or pressure.

\paragraph{Outcome-stratified hazard.}
The left half of Table~\ref{tab:outcome-hazard} reports
mean predicted \emph{Financial Exploitation} hazard@3 by call outcome. A
trajectory-tracking model should rank Both / \emph{Financial
Exploitation}-only above \emph{Information Gathering}-only / Neither.
With text alone, gpt-4.1-mini ZS inverts this ordering
(Neither $>$ \emph{Information Gathering}-only); adding the trajectory
corrects it under both oracle and predicted TTP. The encoder (text+TTP)
shows the cleanest stratification (Both 0.28, \emph{Financial
Exploitation}-only 0.33, \emph{Information Gathering}-only 0.01,
Neither 0.06 with gold; nearly identical with pred), so late-fusion
robustness extends to outcome-conditioned ranking. Inline-fusion SFT
models compress predictions toward zero on non-reaching outcomes,
making low-hazard targets trivial without separating Both /
\emph{Financial Exploitation}-only from the rest.

\paragraph{Post-branch dynamic update.}
The right half of Table~\ref{tab:outcome-hazard}
restricts to Both calls and asks whether a model raises \emph{Financial
Exploitation} hazard \emph{after} the \emph{Information Gathering} branch
is passed. The Markov heuristic is flat ($\Delta = -0.003$ gold, $0.000$
pred): its transition matrix has no notion that ``\emph{Information
Gathering} has occurred'' changes the outlook---the TTP trajectory is
what lets a model update dynamically. gpt-4.1-mini ZS with the trajectory
shows the largest oracle-TTP post-branch lift ($\Delta = +0.231$ gold, $+0.159$
pred), zero-shot GPT-4o holds $\Delta = +0.208$ in both
settings---the largest lift under predicted TTP---and the encoder (text+TTP) keeps $\Delta$ positive in both
settings ($+0.116$ gold, $+0.089$ pred). The late-fusion advantage
therefore carries through to dynamic updating, not only to static
discrimination. Inline-fusion SFT models hold $\Delta\!\approx\!0$ in
both settings: their strong static accuracy on \emph{Financial
Exploitation} comes from a confident static separation of Both calls
from the rest, not from a dynamic re-evaluation at the \emph{Information
Gathering} branch. This dynamic-updating capability
is the \emph{interpretability} contribution of TTP, distinct from raw
accuracy.

\begin{figure*}[!t]
\centering
\includegraphics[width=\textwidth]{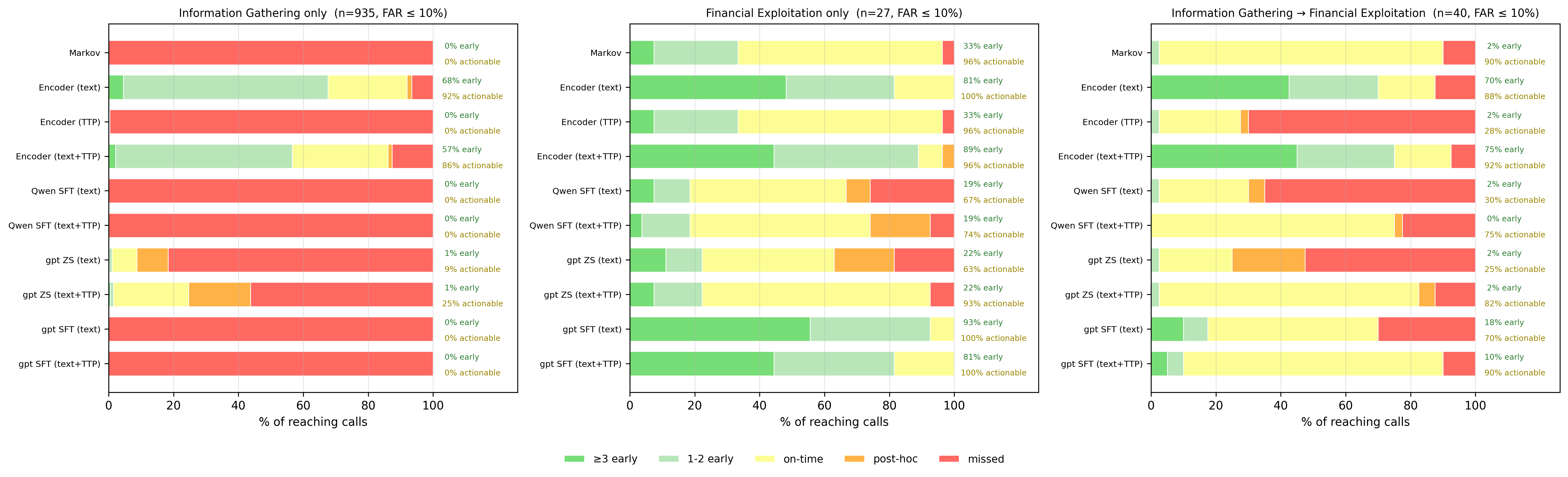}
\caption{Lead-time breakdown at
$\mathrm{FAR}\le 10\%$ under \emph{predicted} TTP, stratified by the
trajectory each call reaches. For each reaching call the lead
$\ell = \tau_g - t_{\text{alarm}}$ is binned into five outcomes; bars
sum to $100\%$ within a (model, stratum) cell. \emph{\% early} counts
$\ell\ge 1$, \emph{\% actionable} counts $\ell\ge 0$.
Stratum sizes: IG only $n{=}935$, FE only $n{=}27$, IG$\to$FE $n{=}40$;
numerical breakdown in Appendix Table~\ref{tab:leadtime-numeric}.
Oracle-TTP version in
Appendix~\ref{sec:appendix-classifier} (Figure~\ref{fig:leadbreak-gold}).}
\label{fig:leadbreak}
\end{figure*}

\paragraph{Lead-time breakdown at a tight false-alarm budget.}
An early-warning alarm must (i)~fire with usable \emph{lead} before the
terminal harm and (ii)~keep its false-alarm rate (FAR) low on calls that
never reach this harm. We cast each model as an alarm firing at the first
turn whose hazard exceeds a threshold $\theta$, fix
$\mathrm{FAR}\le 10\%$ on the non-reaching pool, and bin each
reaching call's lead $\ell = \tau_g - t_{\text{alarm}}$ into
\emph{$\ge\!3$ early} / \emph{$1$--$2$ early} / \emph{on-time} ($\ell{=}0$)
/ \emph{post-hoc} ($\ell{<}0$) / \emph{missed}. We report \emph{\% early}
($\ell\ge 1$, actionable warning at least one turn before harm) and
\emph{\% actionable} ($\ell\ge 0$). Because every test call is a fraud
call (no benign controls), we stratify by trajectory:
\emph{Information Gathering only}, \emph{Financial Exploitation only},
and \emph{Information Gathering $\to$ Financial Exploitation}.
Figure~\ref{fig:leadbreak} reports the breakdown.

Three findings emerge under predicted TTP.
(i)~The TTP signal pays off on the \emph{Information
Gathering $\to$ Financial Exploitation} stratum ($n{=}40$ in test): adding the predicted
trajectory lifts the encoder by 5~pp in \emph{\% early}
($70\%\!\to\!75\%$), because the $T1004$ precursor is the cue the model
leans on to anticipate $T1005$. (ii)~Direct extortion belongs to the
generation-style LLM: fine-tuned \texttt{gpt-4.1-mini} (text) flags $93\%$
of \emph{Financial Exploitation only} calls with $\ell\ge 1$ (FE-only stratum, $n{=}27$, so this percentage is high-variance), since the
abrupt money demand carries surface signal without a precursor; the same
model collapses to $18\%$ early once the call has an \emph{Information
Gathering} preamble. Adding TTP to this inline-fusion model under noisy
predictions degrades both---\emph{FE-only} drops to $82\%$ and the
trajectory stratum to $10\%$---reinforcing the inline-fragility pattern
of \S\ref{sec:results}. (iii)~The Markov baseline fires at or before the
harm turn on $96\%$ of \emph{Financial Exploitation only} calls under
predicted TTP (33\% early + 63\% on-time)---a primarily concurrent
detector. On \emph{Information Gathering only} (86\% base rate), only
the encoder variants remain actionable under $\mathrm{FAR}\le 10\%$.
Appendix~\ref{sec:appendix-qual} walks through one
successful early warning and one missed abrupt demand.

\section{Discussion}
\label{sec:discussion}

\paragraph{From detection to projection.}
We reframe vishing defense as \emph{harm projection} rather than
post-hoc detection. In situation-awareness terms, the question is whether
the model's \emph{tactical comprehension} (the TTP trajectory) is a useful
state for \emph{projecting} which concrete harm a call approaches, so
that intervention is timed and targeted.

\paragraph{The TTP trajectory as an interpretable tactical state.}
The TTP trajectory functions as an \emph{interpretable tactical state}
separate from accuracy gain. On the common, gradual harm, text-only
models already saturate and the trajectory adds little; on the rare,
abrupt harm, the trajectory is the dominant signal for raising
hazard estimates and locating the harm branch in time. It matters
where text alone falls short---telling the model \emph{which} harm a
call is heading toward and \emph{when} it will arrive.

\paragraph{Privacy-conscious deployment.}
A TTP-only encoder makes a privacy-conscious pipeline feasible: an
on-device tactic classifier processes the transcript locally, and only
the abstract tactic sequence is sent to the forecasting service. This
aligns with data-minimization principles at a real accuracy cost relative
to text+TTP; we do not evaluate the deployment empirically but note it
as an architectural option text-conditioned baselines cannot offer.

\paragraph{Transition structure should be learned.}
LLMs given the TTP trajectory still did not match Markov on
\emph{Financial Exploitation} branch timing. Tactic-to-tactic transition
is a corpus-specific statistic that data estimates better than prior
knowledge approximates; a strong ETF model should combine text semantics
with learned transitions.

\section{Conclusion}

We presented Vishing-Tactics-Bench, an utterance-level
benchmark grounded in Endsley's situation-awareness framework (perception,
comprehension, projection). We built a corpus with turn-level TTP labels, defined
the Exploitation Trajectory Forecasting task, and analyzed it with
heuristic and LLM baselines; the tactical trajectory is a key predictive
signal, best exploited when learned from data.

The harm a call inflicts occurs only at its terminal tactics, and
everything before them is a window in which intervention is still
possible. \emph{Which} terminal harm is forecasted determines the
appropriate response (credential-entry blocks and app-installation
warnings for \emph{Information Gathering}, payment blocks and bank
handoff for \emph{Financial Exploitation}), so targeted defense depends
on harm specificity rather than raw detection accuracy. The gap our
baselines leave open is the remaining headroom for practical
early-warning systems.

\paragraph{Future work.}
A next step is to validate ETF in real-time deployment by streaming
audio through an ASR pipeline and measuring lead time and FAR at a
deployed threshold. A further direction is \emph{victim-conditioned}
ETF: victim responses (compliance, resistance, hesitation) shape the
scammer's next tactic, so the trajectory is a function of dialogue
dynamics the current call-as-sequence formulation does not model.
Extending the taxonomy and corpus to other languages and to
authentic-call data is a second direction; cross-lingual transfer of the
trajectory structure is an open question. Alternative hazard
parameterizations (joint hazards, longer horizons,
competing-risks survival architectures) remain future work.

\section*{Limitations}

\paragraph{Synthetic data.}
Real vishing calls are private conversations and crime-related material, and
their collection and release are strictly restricted by legal and
ethical constraints in essentially every jurisdiction. Synthetic
data is therefore a necessary stepping stone for community progress on this domain:
it allows shared, reproducible benchmarks to exist at all, and the
construction of synthetic vishing data has itself become an active research topic.
TeleAntiFraud-28k, on which our corpus is based, was likewise constructed and
released this way. Specifically, TeleAntiFraud-28k publicly releases
only its synthetic subsets; the real-recording subset is not publicly available, which
further constrains the choice of base data. Building on a synthetic base is thus a deliberate
research-design choice, not a substitute we settled for; synthetic conversations
and audio may not fully reflect real calls, and validation on real recordings
remains future work contingent on the resolution of the underlying
legal access barriers.

\paragraph{Single language.}
The corpus is limited to Chinese vishing conversations; generalization to other languages
and cultures is not verified.

\paragraph{Difficulty gap between the two terminal harms.}
The difference in prediction difficulty between \emph{Information Gathering}
and \emph{Financial Exploitation} may stem in part from the difference in their
reach frequency, that is, the amount of positive examples. \emph{Financial
Exploitation} (6.4\%) has far fewer positives than \emph{Information Gathering}
(85.8\%), so it is hard to fully separate whether its low AP reflects intrinsic
difficulty or sample scarcity.

\paragraph{No benign-call controls.}
All calls in the benchmark are vishing calls, so the reported
false-alarm rates count alarms on vishing calls that do not reach the
target harm; false-alarm behaviour on legitimate calls is not measured,
and adding benign-call controls is left for future work.

\section*{Ethics Statement}

\paragraph{Data and privacy.}
All conversations are synthetic (TeleAntiFraud-28k) and contain no real
individuals' personally identifying information. Our
artifact will be released as a gated Hugging Face dataset containing
only the TTP annotations and join keys linked to the Apache-2.0
TeleAntiFraud corpus, so no PII is introduced by our release.

\paragraph{Potential risks.}
Tactic-level annotations of vishing calls could in principle be misused
to craft more deceptive scams. Two factors mitigate this risk: (i) the
annotated corpus is itself synthetic (no real victim utterances) and
already publicly released under Apache-2.0; (ii) the abstract tactic
labels (Trust Building, Pressure, etc.) are common-knowledge
descriptions of social engineering and are not new attack capabilities.
We release the resource to advance defensive harm-projection research,
which is the primary intended use.

\paragraph{AI disclosure.}
Generative AI tools were used to assist with this work: Claude (Anthropic) for
code development and manuscript language editing, and Gemini (Google) for figure
preparation. All AI-generated code was reviewed and tested, and AI-assisted figures
verified, by the authors before use. AI-assisted manuscript edits were limited
to language polishing; all scientific content was written and verified by the
authors.

\section*{Acknowledgments}
This work was supported by the Institute of Information and Communications Technology Planning and Evaluation (IITP) grants funded by the Korea government (MSIT) (No. RS-2025-02215393, Development of Detection and Prediction Technology for New and Unknown Voice Phishing; No. RS-2019-II190004, Development of Semi-supervised Learning Language Intelligence Technology and Korean Tutoring Service for Foreigners).

\bibliography{custom}

\appendix
\input{appendix}

\end{document}

%% file: appendix.tex

\section{Vishing-Tactics Taxonomy and Annotation}
\label{sec:appendix-a}

\subsection{Full Taxonomy: 6 Tactics, 24 Techniques}

The Vishing-Tactics taxonomy defines six tactics divided into 24
techniques (Table~\ref{tab:taxonomy}); counts are over the
Vishing-Tactics-Bench standard split. T1001 (Initial
Contact) is part of the taxonomy but does not apply to this corpus,
which captures only mid-call segments---annotation covers the
remaining five tactics (T1002--T1006) and NONE.

\begin{table}[h]
\centering
\small
\begin{tabular}{llr}
\toprule
\textbf{Technique} & \textbf{Name} & \textbf{Count} \\
\midrule
\multicolumn{3}{l}{\textit{T1001 Initial Contact}} \\
T1001.001 & Phone call & --- \\
T1001.002 & SMS / messaging & --- \\
T1001.003 & Recorded voice / ARS & --- \\
\midrule
\multicolumn{3}{l}{\textit{T1002 Trust Building}} \\
T1002.001 & Impersonating institutions & 5{,}779 \\
T1002.002 & Impersonating acquaintances & 205 \\
T1002.003 & Building rapport & 1{,}558 \\
T1002.004 & Impersonating experts & 820 \\
\midrule
\multicolumn{3}{l}{\textit{T1003 Psychological Pressure}} \\
T1003.001 & Creating urgency & 3{,}332 \\
T1003.002 & Fear / threat & 1{,}202 \\
T1003.003 & Benefit lure & 4{,}512 \\
T1003.004 & Inducing guilt & 16 \\
T1003.005 & Information control & 462 \\
\midrule
\multicolumn{3}{l}{\textit{T1004 Information Gathering}} \\
T1004.001 & PII request & 3{,}455 \\
T1004.002 & Financial info request & 1{,}272 \\
T1004.003 & Malicious app installation & 3{,}651 \\
T1004.004 & Fake website lure & 2{,}603 \\
\midrule
\multicolumn{3}{l}{\textit{T1005 Financial Exploitation}} \\
T1005.001 & Account transfer & 447 \\
T1005.002 & Cash withdrawal & 7 \\
T1005.003 & Voucher / crypto purchase & 2 \\
T1005.004 & Loan exploitation & 10 \\
T1005.005 & Account opening abuse & 3 \\
\midrule
\multicolumn{3}{l}{\textit{T1006 Evasion}} \\
T1006.001 & Conversation deletion & 3 \\
T1006.002 & Disappearing act & 402 \\
T1006.003 & Identity theft / burner phone & 36 \\
\midrule
NONE & (no tactical intent) & 5{,}292 \\
\bottomrule
\end{tabular}
\caption{The Vishing-Tactics taxonomy and technique-level counts over
the Vishing-Tactics-Bench standard split
(T1001 does not apply to mid-call recordings).
All reported experiments use tactic-level labels;
technique-level annotations are released for future fine-grained studies
and are not used for training or evaluation.
271 turns (0.8\%) carry a tactic label without a technique sub-label
and appear only in the tactic-level totals of
Table~\ref{tab:tactic-dist}.}
\label{tab:taxonomy}
\end{table}

Six of the 21 annotated techniques have fewer than 35 instances and four are
in single digits, so the benchmark uses tactic-level labels; technique-level
annotation is released for fine-grained analysis.

\subsection{Annotation Procedure}

Tactic labels were produced by cross-annotation with two large language models
(Claude and GPT-4o). Each model was called once per call, received the full call
context, and independently labeled every scammer utterance with a tactic, technique,
confidence, and reasoning (temperature 0.1). Inter-model agreement was Cohen's
$\kappa = 0.94$. The annotation prompt has the following skeleton (the taxonomy
description expands to the five annotated tactics and 21 techniques):

\begin{prompttemplate}{Annotation prompt}
\small\ttfamily
You are a professional analyst of voice phishing conversations. Analyze the following
Chinese phone conversation and assign the most appropriate TTP label to the scammer's
utterances only. [taxonomy] [conversation] Rules: judge from full context; label a
turn with no clear fraud pattern as NONE; pick one technique per turn; give a
confidence and a brief reasoning. Output strict JSON.
\end{prompttemplate}

\paragraph{Human validation.}
To validate the LLM-based annotation, two human annotators
independently relabeled a stratified random sample of 118 calls (777
scammer turns), about 10\% of the test split. The sample includes all
\emph{Financial Exploitation}-reach calls
(\emph{Information Gathering}$\to$\emph{Financial Exploitation} and
\emph{Financial Exploitation}-only; 67 calls) together with a stratified
draw of \emph{Information Gathering}-only and Neither calls (51 calls),
additionally oversampling \emph{Evasion}-containing calls and lightly
boosting under-represented phishing, so that all six tactic classes
including the rare \emph{Financial Exploitation} and \emph{Evasion} are
well covered. The first call of the sample was used as a
calibration round and is excluded from the agreement computation.
Annotators viewed the full call context (as the LLM
annotators did), since label quality is an offline ground-truth concern,
distinct from the runtime classifier of
Appendix~\ref{sec:appendix-classifier} (which is restricted to prefix-only
input for streaming compatibility).

\paragraph{Inter-annotator agreement.}
Table~\ref{tab:iaa-pairwise} reports pairwise Cohen's
$\kappa$ across the four raters (HA, HB, M1, M2) and the 4-rater Fleiss's
$\kappa$. Human--human agreement is substantial ($\kappa = 0.69$); LLM--LLM
agreement is near-perfect on this subset ($\kappa = 0.97$, marginally
higher than the corpus-level model--model agreement of $\kappa = 0.94$);
4-rater Fleiss's $\kappa = 0.66$. Combined, this supports use of the
LLM-annotated labels as a stand-in for human judgment at corpus scale.
Per-class agreement is strongest on \emph{Information Gathering}
(HA--HB $\kappa_{\text{class}}=0.79$) and \emph{Financial Exploitation}
($\kappa_{\text{class}}=0.76$); residual disagreement concentrates on the
\emph{Trust Building} / \emph{Psychological Pressure} boundary, an
inherently fuzzy distinction in social-engineering speech.
Table~\ref{tab:iaa-perclass} reports per-class
agreement for all rater pairs, and Table~\ref{tab:iaa-confusion} the
pooled human--model label confusion.

\begin{table}[h]
\centering
\small
\begin{tabular}{lc}
\toprule
\textbf{Pair} & \textbf{Cohen's $\kappa$} \\
\midrule
M1 (Claude) vs M2 (GPT-4o) & 0.97 \\
HA vs HB (human--human) & 0.69 \\
HA vs M1 & 0.51 \\
HA vs M2 & 0.52 \\
HB vs M1 & 0.64 \\
HB vs M2 & 0.66 \\
\midrule
4-rater Fleiss's $\kappa$ & 0.66 \\
\bottomrule
\end{tabular}
\caption{Pairwise inter-annotator agreement on the
774-turn human-validation sample (calibration call
excluded). All pairs reach moderate to substantial
agreement; the two LLM annotators are near-perfect.}
\label{tab:iaa-pairwise}
\end{table}

\begin{table}[h]
\centering
\fontsize{8.2}{10}\selectfont
\setlength{\tabcolsep}{2pt}
\begin{tabular}{lcccccc}
\toprule
\textbf{Tactic} & \textbf{HA--HB} & \textbf{HA--M1} & \textbf{HA--M2} & \textbf{HB--M1} & \textbf{HB--M2} & \textbf{M1--M2} \\
\midrule
T1002 & 0.48 & 0.41 & 0.39 & 0.48 & 0.51 & 0.97 \\
T1003 & 0.72 & 0.44 & 0.44 & 0.65 & 0.65 & 0.99 \\
T1004 & 0.79 & 0.67 & 0.68 & 0.82 & 0.81 & 0.98 \\
T1005 & 0.76 & 0.54 & 0.55 & 0.65 & 0.66 & 0.97 \\
T1006 & 0.51 & 0.43 & 0.47 & 0.72 & 0.73 & 0.91 \\
NONE  & 0.77 & 0.53 & 0.55 & 0.60 & 0.67 & 0.91 \\
\bottomrule
\end{tabular}
\caption{Per-class one-vs-rest Cohen's $\kappa$ for all rater pairs on
the 774-turn human-validation sample.}
\label{tab:iaa-perclass}
\end{table}

\begin{table}[h]
\centering
\small
\setlength{\tabcolsep}{4pt}
\begin{tabular}{lrrrrrr}
\toprule
& \multicolumn{6}{c}{\textbf{Model label}} \\
\cmidrule(lr){2-7}
\shortstack[l]{\textbf{Human}\\\textbf{label}} & T1002 & T1003 & T1004 & T1005 & T1006 & NONE \\
\midrule
T1002 & 348 & 57 & 29 & 21 & 29 & 88 \\
T1003 & 185 & 625 & 14 & 19 & 8 & 27 \\
T1004 & 68 & 93 & 560 & 19 & 5 & 19 \\
T1005 & 46 & 166 & 24 & 294 & 2 & 6 \\
T1006 & 9 & 8 & 2 & 1 & 71 & 39 \\
NONE  & 8 & 13 & 7 & 6 & 3 & 177 \\
\bottomrule
\end{tabular}
\caption{Human--model label confusion on the human-validation sample,
pooled over the four human--model rater pairs (HA/HB $\times$ M1/M2).}
\label{tab:iaa-confusion}
\end{table}

\section{Corpus Statistics}
\label{sec:appendix-b}

The statistics below are computed over the full ETF standard split
(train + test, $n = 5{,}645$ calls).

\subsection{Composition and reach by fraud type}

Table~\ref{tab:split} reports the standard split by fraud type.
Table~\ref{tab:reach} reports call-level reach rates of the two terminal
harms by fraud type.

\begin{table}[h]
\centering
\small
\begin{tabular}{llrrr}
\toprule
\textbf{Fraud type} & \textbf{Unit} & \textbf{Train} & \textbf{Test} & \textbf{Total} \\
\midrule
Customer-service & calls & 1,564 & 391 & 1,955 \\
                 & turns & 9,626 & 2,383 & 12,009 \\
Bank             & calls & 1,426 & 357 & 1,783 \\
                 & turns & 9,125 & 2,330 & 11,455 \\
Investment       & calls & 643 & 161 & 804 \\
                 & turns & 4,720 & 1,170 & 5,890 \\
Phishing         & calls & 458 & 115 & 573 \\
                 & turns & 2,937 & 741 & 3,678 \\
Lottery          & calls & 240 & 60 & 300 \\
                 & turns & 1,276 & 325 & 1,601 \\
Kidnap           & calls & 184 & 46 & 230 \\
                 & turns & 538 & 169 & 707 \\
\midrule
\textbf{Total}   & calls & \textbf{4,515} & \textbf{1,130} & \textbf{5,645} \\
                 & turns & \textbf{28,222} & \textbf{7,118} & \textbf{35,340} \\
\bottomrule
\end{tabular}
\caption{Vishing-Tactics-Bench standard split. Each fraud type is reported on two rows: number
of calls and number of scammer turns.}
\label{tab:split}
\end{table}

\begin{table}[h]
\centering
\small
\begin{tabular}{lrr}
\toprule
\textbf{Fraud type} & \textbf{\emph{IG} (\%)} & \textbf{\emph{FE} (\%)} \\
\midrule
Customer-service & 88.9 & 6.0 \\
Bank             & 93.4 & 3.0 \\
Investment       & 81.5 & 11.4 \\
Phishing         & 98.6 & 0.7 \\
Lottery          & 69.0 & 10.3 \\
Kidnap           & 5.7 & 28.7 \\
\midrule
\textbf{Overall} & \textbf{85.8} & \textbf{6.4} \\
\bottomrule
\end{tabular}
\caption{Reach rate of the two terminal harms (\emph{Information Gathering} and \emph{Financial Exploitation}) by fraud type (call-level).}
\label{tab:reach}
\end{table}

\subsection{Co-occurrence and timing of the Two Terminal Harms}

Table~\ref{tab:quadrant} groups calls by which terminal harms they reach.
Of the 365 calls that reach financial exploitation, 232 (64\%) pass
through information gathering and 133 (36\%) reach it directly; of the
calls that reach information gathering, only about 5\% go on to financial
exploitation. The two terminal harms are thus in a partial
relation---neither a simple escalation ladder nor fully independent---and
this relation varies by fraud type. When financial exploitation is
reached, the first \emph{Financial Exploitation} utterance occurs at a mean of 4.0 scammer turns
(median 4, std 2.4, range 1--17), about 65\% of the way through the call,
which is why its branch point is captured sharply in the body
(Section~\ref{sec:results}).

\begin{table}[h]
\centering
\small
\begin{tabular}{lrr}
\toprule
\textbf{Reach pattern} & \textbf{Calls} & \textbf{\%} \\
\midrule
Both \emph{IG} and \emph{FE} & 232 & 4.1 \\
\emph{IG} only & 4{,}611 & 81.7 \\
\emph{FE} only (no \emph{IG}) & 133 & 2.4 \\
Neither & 669 & 11.9 \\
\bottomrule
\end{tabular}
\caption{Co-occurrence of the two terminal harms (\emph{Information Gathering} and \emph{Financial Exploitation}) at the call level.}
\label{tab:quadrant}
\end{table}

\subsection{Per-row uncertainty for Table~\ref{tab:results}}
\label{sec:uncertainty}

Tables~\ref{tab:results-uncertainty-t1004} and
\ref{tab:results-uncertainty-t1005} report per-row uncertainty for the
main results table, split by terminal harm. Encoder (3 variants) and
Qwen2.5-7B SFT (2 variants) report mean $\pm$ std across 3 seeds (42,
1337, 2024); Markov, ZS LLMs, and gpt-4.1-mini SFT are single runs whose
uncertainty is reported via the 95\% bootstrap CI over turn-level pairs
($B = 2000$), since Markov is deterministic, the OpenAI tuning API does
not expose seeded retraining, and ZS LLMs are queried at temperature
$0.1$.

\begin{table*}[h]
\centering
\small
\setlength{\tabcolsep}{6pt}
\begin{tabular}{lccc}
\toprule
\textbf{Model} & \textbf{AP@1\,/\,3\,/\,5}\,$\uparrow$ & \textbf{C-idx}\,$\uparrow$ & \textbf{Div.\,err.}\,$\downarrow$ \\
\midrule
Markov heuristic            & .40[.38,.42] / .49[.48,.50] / .52[.50,.53] & .55[.54,.57] & 1.45[1.35,1.55] \\
\midrule
Encoder (text)              & .89$\pm$.00 / .95$\pm$.00 / .97$\pm$.00 & .98$\pm$.00 & 1.43$\pm$.18 \\
Encoder (TTP)               & .32$\pm$.01 / .71$\pm$.01 / .79$\pm$.00 & .84$\pm$.01 & 2.10$\pm$.05 \\
Encoder (text+TTP)          & .89$\pm$.01 / .95$\pm$.00 / .97$\pm$.00 & .98$\pm$.00 & 1.36$\pm$.07 \\
\midrule
Qwen2.5-7B SFT (text)       & .54$\pm$.01 / .73$\pm$.00 / .80$\pm$.00 & .86$\pm$.00 & 1.51$\pm$.04 \\
Qwen2.5-7B SFT (text+TTP)   & .67$\pm$.01 / .81$\pm$.00 / .88$\pm$.00 & .93$\pm$.00 & 1.48$\pm$.02 \\
\midrule
gpt-4.1-mini ZS (text)      & .27[.26,.28] / .39[.38,.40] / .41[.40,.42] & .39[.37,.40] & 1.22[1.13,1.32] \\
gpt-4.1-mini ZS (text+TTP)  & .36[.34,.38] / .43[.41,.44] / .45[.43,.46] & .45[.44,.47] & 0.58[.51,.66] \\
gpt-4.1-mini SFT (text)     & .87[.85,.88] / .91[.90,.92] / .93[.92,.94] & .96[.95,.96] & 1.24[1.17,1.32] \\
gpt-4.1-mini SFT (text+TTP) & .93[.92,.93] / .96[.95,.96] / .97[.96,.97] & .98[.98,.98] & 1.28[1.19,1.36] \\
\bottomrule
\end{tabular}
\caption{Per-row uncertainty for the main ETF results
(Table~\ref{tab:results})---\textbf{\emph{Information Gathering}} only.
Conventions as described in this subsection; leading zeros on AP@$k$ and
C-index are dropped to save space.}
\label{tab:results-uncertainty-t1004}
\end{table*}

\begin{table*}[h]
\centering
\small
\setlength{\tabcolsep}{6pt}
\begin{tabular}{lccc}
\toprule
\textbf{Model} & \textbf{AP@1\,/\,3\,/\,5}\,$\uparrow$ & \textbf{C-idx}\,$\uparrow$ & \textbf{Div.\,err.}\,$\downarrow$ \\
\midrule
Markov heuristic            & .39[.30,.47] / .24[.19,.30] / .22[.17,.26] & .67[.64,.70] & 0.07[.00,.19] \\
\midrule
Encoder (text)              & .58$\pm$.02 / .56$\pm$.02 / .56$\pm$.01 & .92$\pm$.01 & 1.15$\pm$.19 \\
Encoder (TTP)               & .11$\pm$.07 / .10$\pm$.04 / .10$\pm$.03 & .68$\pm$.01 & 0.89$\pm$.15 \\
Encoder (text+TTP)          & .59$\pm$.03 / .57$\pm$.04 / .56$\pm$.05 & .91$\pm$.01 & 1.15$\pm$.24 \\
\midrule
Qwen2.5-7B SFT (text)       & .09$\pm$.02 / .08$\pm$.01 / .08$\pm$.01 & .54$\pm$.01 & 1.96$\pm$.22 \\
Qwen2.5-7B SFT (text+TTP)   & .32$\pm$.02 / .20$\pm$.01 / .18$\pm$.01 & .60$\pm$.00 & 0.18$\pm$.08 \\
\midrule
gpt-4.1-mini ZS (text)      & .07[.05,.11] / .09[.07,.12] / .09[.07,.12] & .65[.62,.68] & 1.49[1.07,1.94] \\
gpt-4.1-mini ZS (text+TTP)  & .35[.27,.44] / .22[.17,.27] / .18[.14,.23] & .65[.61,.68] & 0.34[.14,.60] \\
gpt-4.1-mini SFT (text)     & .51[.42,.59] / .46[.39,.52] / .43[.38,.50] & .83[.80,.86] & 0.76[.45,1.14] \\
gpt-4.1-mini SFT (text+TTP) & .67[.59,.74] / .55[.49,.61] / .53[.48,.59] & .86[.83,.88] & 0.00[.00,.00] \\
\bottomrule
\end{tabular}
\caption{Per-row uncertainty for the main ETF results
(Table~\ref{tab:results})---\textbf{\emph{Financial Exploitation}} only.
Conventions match Table~\ref{tab:results-uncertainty-t1004}.}
\label{tab:results-uncertainty-t1005}
\end{table*}

\subsection{Per-fraud-type heterogeneity}
\label{sec:per-fraud-full}

The pooled view of Table~\ref{tab:results} averages across six fraud
types whose trajectory profiles differ substantially: PII-dominant types
(customer service, bank, phishing) account for 76\% of test calls;
investment and lottery (20\%) form a mixed regime; kidnap (4\%) is
finance-direct. The TTP gain is therefore \emph{type-conditional}. On
kidnap, where the \emph{Information Gathering} base rate is an order of
magnitude below other types, adding TTP raises gpt-4.1-mini SFT
\emph{Information Gathering} AP@1 from $0.32$ to $0.77$ ($+0.45$)---the
single largest TTP gain in the benchmark. On investment, whose
trajectories mimic PII-only types before a late branch, the same move
raises \emph{Financial Exploitation} AP@1 by $+0.31$. Conversely, on
bank and lottery, where \emph{Financial Exploitation} phrasing is highly
stereotyped, the TTP gain on \emph{Financial Exploitation} AP@1 is
slightly negative ($-0.03$ on both): text alone already captures the
demand wording. The strongest pooled model is also not the strongest per
type---on bank \emph{Financial Exploitation}, the Encoder with text alone
outranks gpt-4.1-mini SFT (text+TTP) by $+0.26$ AP@1 despite far fewer
parameters and no TTP signal. The TTP gain thus concentrates where the
trajectory shape deviates from the population average and becomes
redundant where harm-bearing phrasing is stereotyped.

\begin{figure*}[!t]
\centering
\includegraphics[width=\textwidth]{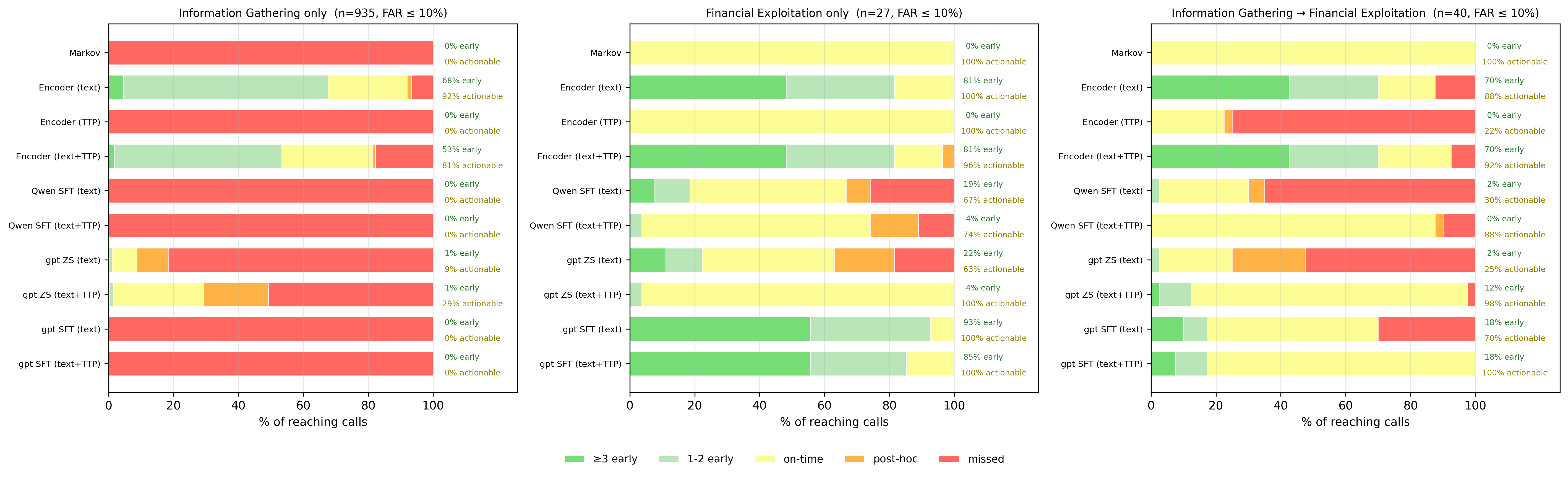}
\caption{Lead-time breakdown at $\mathrm{FAR}\le 10\%$
under \emph{oracle} TTP; conventions follow Figure~\ref{fig:leadbreak}.}
\label{fig:leadbreak-gold}
\end{figure*}

\section{Cascaded TTP Classifier}
\label{sec:appendix-classifier}

Predicted-TTP rows in Tables~\ref{tab:results} and
\ref{tab:outcome-hazard} use a turn-level tactic classifier applied
prefix-only over the standard test split (4{,}515 training calls /
28{,}222 scammer turns; 1{,}130 test calls / 7{,}118 turns). The backbone
is the Chinese RoBERTa encoder \texttt{hfl/chinese-roberta-wwm-ext} (110M
parameters) with a single linear classification head over six classes
($T1002$, $T1003$, $T1004$, $T1005$, $T1006$, NONE). Optimisation is
standard supervised cross-entropy with class weights $w_c \propto
1/(n_c+10)$ to mitigate the rarity of \emph{Financial Exploitation} and
\emph{Evasion} (1.4\% and 1.4\% of training turns), AdamW at learning
rate $2\times10^{-5}$, batch
size 32, three epochs, and a maximum input length of 256 wordpieces. At
inference the model is restricted to the prefix up to and including the
current scammer turn, consistent with a streaming-deployment scenario.

\paragraph{Context-window ablation.}
We compare three context windows fed to the same RoBERTa
backbone under identical optimisation (Table~\ref{tab:classifier-ctx}).
The current scammer turn alone yields the strongest test macro-F1
(0.748); extending to the four preceding turns reduces macro-F1 to
0.701, and the full prefix reduces it further to 0.634. The same ordering
holds across all four non-NONE tactic classes. We attribute this to the
locality of tactic indicators: each TTP class is most reliably signalled
by the lexical surface of the current scammer utterance (e.g., explicit
account or credential requests for \emph{Information Gathering}; payment
or transfer demands for \emph{Financial Exploitation}). Conditioning on
prior turns admixes earlier tactic content into the representation and
weakens the discriminative signal of the current turn. We therefore adopt
the current-turn classifier as the cascaded TTP source for the
predicted-TTP rows in the body tables.

\begin{table}[h!]
\centering
\small
\begin{tabular}{lcc}
\toprule
\textbf{Context} & \textbf{Acc.} & \textbf{Macro-F1}  \\
\midrule
1 turn (current) & \textbf{0.846} & \textbf{0.748}  \\
5 turns & 0.787 & 0.701  \\
All prior turns & 0.758 & 0.634  \\
\bottomrule
\end{tabular}
\caption{Cascaded TTP classifier performance as a function
of the input context window.}
\label{tab:classifier-ctx}
\end{table}

\paragraph{Per-class performance of the current-turn classifier.}
The chosen classifier attains test accuracy 0.846 and macro-F1 0.748,
with per-class F1 of $T1002$ (Trust Building) 0.865, $T1003$
(Psychological Pressure) 0.868, $T1004$ (Information Gathering) 0.903,
$T1005$ (Financial Exploitation) 0.584, $T1006$ (Evasion) 0.517, and
NONE 0.751. The residual error
concentrates on the two rare tactics \emph{Financial Exploitation} and
\emph{Evasion}, which accounts for the asymmetric degradation observed
in Table~\ref{tab:results}: predicted-TTP \emph{Financial Exploitation}
AP@1 drops more sharply than \emph{Information Gathering} AP@1,
mirroring the per-class F1 gap.

\paragraph{Oracle-TTP lead-time breakdown.}
Figure~\ref{fig:leadbreak-gold} reports the lead-time
breakdown under oracle TTP for comparison with the
predicted-TTP figure in the body (Figure~\ref{fig:leadbreak}). The oracle
case shows the upper bound on TTP-driven lift; predicted TTP preserves
the encoder's lead on the \emph{Information Gathering $\to$ Financial
Exploitation} stratum and degrades inline-fusion variants on the
\emph{Financial Exploitation only} stratum, tracking the per-class F1
gap.

\paragraph{Numerical lead-time breakdown.}
Table~\ref{tab:leadtime-numeric} reports the predicted-TTP lead-time
breakdown numerically. The threshold $\theta$ is chosen per (model, goal)
to satisfy $\mathrm{FAR}\le 10\%$ on the non-reaching pool; \emph{\% early}
counts $\ell\ge 1$ and \emph{missed} counts calls that never fire by the
harm turn. Stratum sizes are IG only $n{=}935$, FE only $n{=}27$,
IG$\to$FE $n{=}40$.

\begin{table*}[!t]
\centering
\small
\setlength{\tabcolsep}{4pt}
\begin{tabular}{lrrrrrr}
\toprule
& \multicolumn{2}{c}{\textbf{IG only ($n{=}935$)}}
& \multicolumn{2}{c}{\textbf{FE only ($n{=}27$)}}
& \multicolumn{2}{c}{\textbf{IG$\to$FE ($n{=}40$)}} \\
\cmidrule(lr){2-3}\cmidrule(lr){4-5}\cmidrule(lr){6-7}
\textbf{Model} & \textbf{\% early} & \textbf{missed}
& \textbf{\% early} & \textbf{missed}
& \textbf{\% early} & \textbf{missed} \\
\midrule
Markov heuristic           &  0.0 & 100.0 & 33.3 &  3.7 &  2.5 & 10.0 \\
Encoder (text)             & 67.6 &   6.6 & 81.5 &  0.0 & 70.0 & 12.5 \\
Encoder (TTP)              &  0.2 &  99.7 & 33.3 &  3.7 &  2.5 & 70.0 \\
Encoder (text+TTP)         & 56.6 &  12.6 & 88.9 &  0.0 & 75.0 &  7.5 \\
Qwen2.5-7B SFT (text)      &  0.0 & 100.0 & 18.5 & 25.9 &  2.5 & 65.0 \\
Qwen2.5-7B SFT (text+TTP)  &  0.0 & 100.0 & 18.5 &  7.4 &  0.0 & 22.5 \\
gpt-4.1-mini ZS (text)     &  1.1 &  81.7 & 22.2 & 18.5 &  2.5 & 52.5 \\
gpt-4.1-mini ZS (text+TTP) &  1.5 &  56.3 & 22.2 &  7.4 &  2.5 & 12.5 \\
gpt-4.1-mini SFT (text)    &  0.0 & 100.0 & 92.6 &  0.0 & 17.5 & 30.0 \\
gpt-4.1-mini SFT (text+TTP)&  0.0 & 100.0 & 81.5 &  0.0 & 10.0 & 10.0 \\
\bottomrule
\end{tabular}
\caption{Per-stratum lead-time numerics under
\emph{predicted} TTP at $\mathrm{FAR}\le 10\%$, complementing
Figure~\ref{fig:leadbreak}. \emph{\% early} counts $\ell\ge 1$
(actionable warning at least one turn before the harm); \emph{missed}
counts calls whose alarm never fires by the harm turn.}
\label{tab:leadtime-numeric}
\end{table*}

\section{Sequence-Encoder Ablation}
\label{sec:appendix-seqenc}

Table~\ref{tab:seqenc} replaces the GRU trajectory encoder of the
encoder baselines with an LSTM and a two-layer Transformer under an
identical protocol (embedding dim 32, hidden dim 64, mean over 3 seeds,
gold and predicted TTPs). Late-fusion text+TTP results are within one
seed's variance across the three encoders, and the gold-to-predicted
change in \emph{Financial Exploitation} AP@1 stays within 0.01 for each;
the robustness of late fusion to predicted-TTP noise is a property of
the fusion architecture rather than of the specific sequence encoder.
In the TTP-only setting the Transformer extracts substantially more
signal (\emph{Financial Exploitation} AP@1 0.393 vs.\ 0.108 for the GRU
with gold TTP); with text present this advantage is absorbed by the
saturated text branch. On \emph{Information Gathering} the
Transformer's TTP-only C-index reaches 0.94 under predicted TTP,
narrowing the text-free cost of the privacy-conscious pipeline of
Section~\ref{sec:discussion}, while \emph{Financial Exploitation}
still favors text+TTP.

\begin{table*}[h]
\centering
\small
\setlength{\tabcolsep}{4.6pt}
\begin{tabular}{llcccccc}
\toprule
& & \multicolumn{3}{c}{\textbf{\emph{Information Gathering}}}
& \multicolumn{3}{c}{\textbf{\emph{Financial Exploitation}}} \\
\cmidrule(lr){3-5}\cmidrule(lr){6-8}
\textbf{Encoder} & \textbf{TTP} & \textbf{AP@1/3/5}\,$\uparrow$ & \textbf{C-idx}\,$\uparrow$
& \textbf{Div.\,err.}\,$\downarrow$ & \textbf{AP@1/3/5}\,$\uparrow$
& \textbf{C-idx}\,$\uparrow$ & \textbf{Div.\,err.}\,$\downarrow$ \\
\midrule
\multirow{2}{*}{GRU (TTP)}         & gold & 0.318 / 0.707 / 0.787 & 0.841 & 2.10 & 0.108 / 0.098 / 0.099 & 0.680 & 0.89 \\
                                   & pred & 0.316 / 0.693 / 0.773 & 0.828 & 2.12 & 0.080 / 0.091 / 0.095 & 0.684 & 1.45 \\
\multirow{2}{*}{GRU (text+TTP)}    & gold & 0.889 / 0.950 / 0.968 & 0.976 & 1.36 & 0.593 / 0.571 / 0.559 & 0.915 & 1.15 \\
                                   & pred & 0.889 / 0.950 / 0.968 & 0.976 & 1.38 & 0.597 / 0.578 / 0.567 & 0.918 & 1.25 \\
\midrule
\multirow{2}{*}{LSTM (TTP)}        & gold & 0.352 / 0.636 / 0.732 & 0.803 & 2.34 & 0.086 / 0.081 / 0.088 & 0.638 & 0.69 \\
                                   & pred & 0.348 / 0.610 / 0.701 & 0.779 & 2.35 & 0.066 / 0.074 / 0.082 & 0.647 & 0.95 \\
\multirow{2}{*}{LSTM (text+TTP)}   & gold & 0.892 / 0.949 / 0.966 & 0.974 & 1.40 & 0.578 / 0.560 / 0.545 & 0.909 & 1.09 \\
                                   & pred & 0.893 / 0.950 / 0.967 & 0.975 & 1.37 & 0.577 / 0.563 / 0.551 & 0.908 & 1.26 \\
\midrule
\multirow{2}{*}{Transformer (TTP)} & gold & 0.846 / 0.914 / 0.939 & 0.959 & 1.12 & 0.393 / 0.269 / 0.241 & 0.726 & 0.05 \\
                                   & pred & 0.776 / 0.880 / 0.910 & 0.936 & 1.09 & 0.214 / 0.182 / 0.171 & 0.698 & 0.20 \\
\multirow{2}{*}{Transformer (text+TTP)} & gold & 0.898 / 0.950 / 0.966 & 0.975 & 1.44 & 0.601 / 0.602 / 0.612 & 0.916 & 1.43 \\
                                   & pred & 0.899 / 0.950 / 0.964 & 0.975 & 1.44 & 0.600 / 0.600 / 0.608 & 0.914 & 1.37 \\
\bottomrule
\end{tabular}
\caption{ETF results with alternative sequence encoders over the TTP
trajectory, mean across 3 seeds; the GRU rows correspond to the encoder
rows of Table~\ref{tab:results}. Columns follow
Table~\ref{tab:results}.}
\label{tab:seqenc}
\end{table*}

\section{Qualitative Examples}
\label{sec:appendix-qual}

Two test calls illustrate when the TTP trajectory helps and when it does
not. Bracketed values give the \emph{Financial Exploitation} hazard@3 of
the encoder at each scammer turn
(text-only\,/\,TTP-only\,/\,text+TTP, predicted TTP); both dialogues
are translated from Chinese.
In Example~1 (customer-service impersonation), the
text-only and TTP-only branches stay near zero until the demand is
spoken, while the fused encoder raises the hazard to 0.99 at the first
\emph{Information Gathering} turn, two turns before the harm.
In Example~2 (kidnap fraud), with no
trust-building, pressure, or information-gathering trajectory, the
demand arrives at the second scammer turn: the fused encoder misses it
(0.04) while the text-only encoder reacts to the surface wording of the
wire demand (0.66) --- the pattern of Section~\ref{sec:anatomy}, where
abrupt \emph{Financial Exploitation} carries surface signal but no
trajectory precursor.

\begin{examplebluebox}{Example 1: Early warning from the trajectory --- customer-service impersonation}
\small
\textbf{Scammer} [T1002; 0.02\,/\,0.13\,/\,0.30]: ``Hello, this is Tianlan Tech,
agent Li from customer service. How has your recent shopping experience
on our platform been?''\\
\textbf{Victim}: ``It was fine, no particular problems.''\\
\textbf{Scammer} [T1003; 0.02\,/\,0.02\,/\,0.66]: ``Great --- we value every
customer's experience. By the way, we are running a special promotion:
just download our app and you receive a 100-yuan coupon for free.''\\
\textbf{Victim}: ``Oh really? Where do I download it?''\\
\textbf{Scammer} [T1004; 0.02\,/\,0.02\,/\,\textbf{0.99}]: ``Very simple --- just
click this link and the app downloads directly.''\\
\textbf{Scammer} [T1004; 0.02\,/\,0.02\,/\,0.85]: ``After downloading, register
and enter the invitation code; the coupon is credited automatically.''\\
\textbf{Scammer} [T1005 reached; 0.99\,/\,0.02\,/\,1.00]: ``And to keep your
account safe, we release the coupon after you complete one small
transaction --- is that convenient?''\\
\textbf{Victim}: ``How much is the small transaction?''\\
\textbf{Scammer} [T1005]: ``Only 5 yuan, just to verify that your
account is genuine.''
\end{examplebluebox}

\begin{exampleredbox}{Example 2: Missed abrupt demand --- kidnap fraud}
\small
\textbf{Scammer} [T1002; 0.05\,/\,0.13\,/\,0.00]: ``Hello, this is the
international finance center. We regret to inform you that your son ran
into trouble while traveling abroad. He is being held by a local gang
--- you must wire 1,000,000 RMB immediately as ransom.''\\
\textbf{Scammer} [T1005 reached; 0.66\,/\,0.39\,/\,\textbf{0.04}]: ``Otherwise
they may hurt him. Please transfer the money to the designated account
at once.''\\
\textbf{Scammer} [T1005; 0.94\,/\,0.40\,/\,0.93]: ``I will send you the account
details.''\\
\textbf{Scammer} [T1003; 0.05\,/\,0.31\,/\,0.01]: ``Time is extremely tight ---
you must act now.''\\
\textbf{Victim}: ``Wait --- if my son is really being held abroad, I
need concrete proof, such as a photo or a video.''
\end{exampleredbox}